\documentclass{article}
\usepackage{ijcai24}

\usepackage{times}
\usepackage{soul}
\usepackage{url}
\usepackage[hidelinks]{hyperref}
\usepackage[utf8]{inputenc}
\usepackage[small]{caption}
\usepackage{graphicx}
\usepackage{amsmath}
\usepackage{amsthm}
\usepackage{booktabs}
\usepackage{algorithm}
\usepackage{algorithmic}
\usepackage[switch]{lineno}
\usepackage{amssymb}
\usepackage{multirow}
\title{Cross-Domain Few-Shot Semantic Segmentation via Doubly Matching Transformation}

\author{
Jiayi Chen$^{1,2}$\and
Rong Quan$^{1,2}$\and
Jie Qin$^{1,2,*}$\\
\affiliations
$^1$Nanjing University of Aeronautics and Astronautics\\
$^2$State Key Laboratory of Integrated Services Networks, Xidian University\\
\emails
chenjiayi68@nuaa.edu.cn,
\{rongquan0806, qinjiebuaa\}@gmail.com
}
\begin{document}

\maketitle

\begin{abstract}
    Cross-Domain Few-shot Semantic Segmentation (CD-FSS) aims to train generalized models that can segment classes from different domains with a few labeled images. Previous works have proven the effectiveness of feature transformation in addressing CD-FSS. However, they completely rely on support images for feature transformation, and repeatedly utilizing a few support images for each class may easily lead to overfitting and overlooking intra-class appearance differences. In this paper, we propose a Doubly Matching Transformation-based Network (DMTNet) to solve the above issue. Instead of completely relying on support images, we propose Self-Matching Transformation (SMT) to construct query-specific transformation matrices based on query images themselves to transform domain-specific query features into domain-agnostic ones. Calculating query-specific transformation matrices can prevent overfitting, especially for the meta-testing stage where only one or several images are used as support images to segment hundreds or thousands of images. After obtaining domain-agnostic features, we exploit a Dual Hypercorrelation Construction (DHC) module to explore the hypercorrelations between the query image with the foreground and background of the support image, based on which foreground and background prediction maps are generated and supervised, respectively, to enhance the segmentation result. In addition, we propose a Test-time Self-Finetuning (TSF) strategy to more accurately self-tune the query prediction in unseen domains. Extensive experiments on four popular datasets show that DMTNet achieves superior performance over state-of-the-art approaches. Code is available at \href{https://github.com/ChenJiayi68/DMTNet}{https://github.com/ChenJiayi68/DMTNet}.

\end{abstract}

\renewcommand{\thefootnote}{}
\footnotetext{* Corresponding Author}

\section{Introduction}
Relying on large-scale labeled datasets~\cite{Ros2016TheSD,Richter2016PlayingFD,Silberman2012IndoorSA}, semantic segmentation~\cite{long2015fully,Zhao2016PyramidSP,chen2017deeplab,Xie2021SegFormerSA} has achieved rapid development in recent years. However, it is difficult to collect such a large amount of training data which requires massive time and expensive annotation costs in practical scenarios. Few-shot Semantic Segmentation (FSS)~\cite{Shaban2017OneShotLF} has been proposed to reduce the heavy dependence of traditional semantic segmentation models on a large number of labeled images. FSS aims to achieve accurate segmentation of a query image only using a few annotated support images. Existing FSS methods~\cite{Rakelly2018ConditionalNF,Zhang2018SGOneSG,Wang2019PANetFI,Liu2020CRNetCN,Okazawa2022InterclassPR} usually adopt meta-learning~\cite{Vinyals2016MatchingNF,Snell2017PrototypicalNF}, which consists of two stages: meta-training and meta-testing. In the meta-training stage, an FSS model is trained on many meta-tasks using base classes. The trained model can then perform accurate segmentation on novel classes in the meta-testing stage.

However, in practical applications, there always exists a large domain gap between source and target datasets due to different label spaces and feature distributions, causing inferior generalization of FSS models to unseen domains. 
To solve the problem of significant performance degradation of FSS models under cross-domain scenarios, Cross-Domain Few-shot Semantic Segmentation (CD-FSS)~\cite{Lei2022CrossDomainFS} is proposed to simultaneously solve the problems of few shot and domain gaps. 
The primary CD-FSS method, PATNet~\cite{Lei2022CrossDomainFS}, eliminates the domain gap by transforming domain-specific features into domain-agnostic ones and conducting segmentation in the domain-agnostic feature space. It combines the prototype set of support images with some learned anchor layers to construct transformation matrices, based on which domain-specific features are transformed into domain-agnostic ones. However, during the meta-testing stage, generating domain-agnostic features for hundreds or thousands of query images just based on the transformation matrices obtained from one or several support images can easily lead to overfitting. In addition, we find that most existing CD-FSS methods~\cite{Min2021HypercorrelationSF,Lei2022CrossDomainFS} only concentrate on foreground object regions and ignore background regions during the segmentation process. They directly filter out the background of support images and only construct dense correlations with foreground objects.
Considering that objects belonging to the same class mostly lie in similar environments, segmentation based on the similarities between not only the foreground objects but also the backgrounds is very likely to result in better performance. 

Based on the above considerations, we propose a novel Doubly Matching Transformation-based Network (DMTNet) for cross-domain few-shot semantic segmentation. DMTNet first exploits a Self-Matching Transformation (SMT) module to construct a unique transformation matrix for each image based on its own prototype. Then, domain-specific features of each query and support image are transformed into domain-agnostic ones self-adaptively, which can avoid overfitting during the meta-testing stage. After obtaining domain-agnostic features, we propose a Dual Hypercorrelation Construction (DHC) module to learn the hypercorrelation between the query image with both the foreground and background of the support image, and generate foreground and background predictions, correspondingly. Supervising both object-wise and background-wise segmentation can enhance the training performance considering the possible similarities between the backgrounds of similar objects. In addition, in the meta-testing stage, we design a Test-time Self-Finetuning (TSF) strategy to further improve query predictions in unseen domains by self-tuning a handful of parameters in the network with support images.

To evaluate the performance of DMTNet, we conduct extensive experiments and ablation studies on four benchmark datasets, including ISIC2018~\cite{Codella2019SkinLA}, Chest X-ray~\cite{Candemir2014LungSI}, Deepglobe~\cite{Demir2018DeepGlobe2A}, and FSS-1000~\cite{Wei2019FSS1000A1}. Experimental results show that DMTNet achieves remarkable improvement, surpassing state-of-the-art approaches on all four datasets.

In summary, our main contributions are as follows:
\begin{itemize}
    \item We propose a novel Doubly Matching Transformation-based Network (DMTNet) for CD-FSS. Instead of completely relying on support images, we propose a Self-Matching Transformation (SMT) module to transform domain-specific query features into domain-agnostic ones in a self-adaptive manner. Compared with transformation completely based on one or several support images, DMTNet can avoid overfitting to a large extent.
    \item 
    We propose a Dual Hypercorrelation Construction (DHC) module to learn the hypercorrelations between the query image and both the foreground and background of the support image. For the first time, we execute segmentation based on both foreground object similarities and background similarities.
    \item We design a Test-time Self-Finetuning (TSF) strategy for meta-testing. Only self-tuning a handful of parameters in the network can significantly improve query predictions in target domains while maintaining minimal complexity.
    \item Extensive experimental results show that DMTNet achieves state-of-the-art performance on four CD-FSS benchmarks, \emph{i.e.}, ISIC2018, Chest X-ray, Deepglobe, and FSS-1000.
\end{itemize}
\begin{figure*}[ht]
\centering
\includegraphics[scale=0.5]{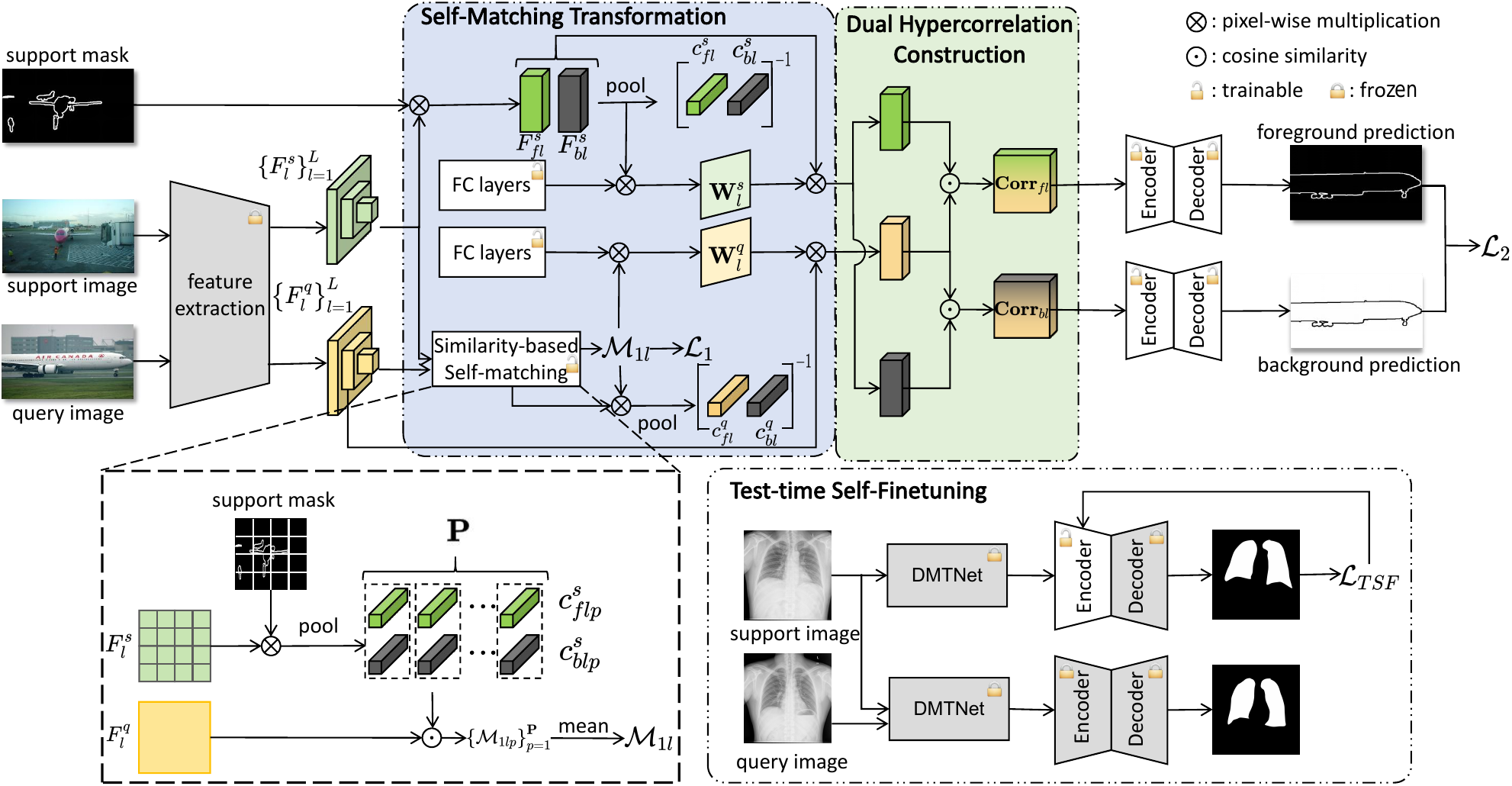}
\caption{Overall architecture of the proposed DMTNet. After obtaining the pyramid features of support and query images, Self-Matching Transformation module (SMT) learns each image a self-adaptive transformation matrix, to transform its domain-specific features into domain-agnostic ones. Then, the Dual Hypercorrelation Construction (DHC) module is introduced to construct dense correlations between the query image with both the foreground and background of the support image. In the meta-testing stage, the Test-time Self-Finetuning (TSF) strategy fine-tunes a few parameters of the encoder to further improve the segmentation performance.}
\label{fig2}
\end{figure*}
\section{Related Work}

\textbf{Few-shot Semantic Segmentation.} FSS aims to generate a pixel-wise prediction of the novel class with only a few labeled support images. Existing FSS methods can be divided into two categories: metric-based methods and relation-based ones. Metric-based methods~\cite{Wang2019PANetFI,Liu2020PartawarePN} represent support images as several class prototypes by utilizing masked average pooling, use non-parametric measurement tools such as cosine similarity to measure the similarity between these prototypes and query features, and segment the query image based on the similarities. However, metric-based methods lose much spatial information when compressing the global feature maps into a prototype vector~\cite{Li2021AdaptivePL,Zhang2021FewShotSV}. Since the prototype has a limited ability to express a target category, some researchers propose relation-based methods~\cite{Zhang2019CANetCS,Yang2020PrototypeMM,Tian2020PriorGF,Min2021HypercorrelationSF} to construct dense correspondences between support-query pairs by calculating their similarities. However, the segmentation performance of these FSS methods will decrease rapidly when facing a large domain gap between the source and the target domain and cannot generalize well to the target domain.\\
\noindent\textbf{Cross-domain Semantic Segmentation.} Cross-domain semantic segmentation can be categorized into domain adaptive semantic segmentation (DASS) and domain generalized semantic segmentation (DGSS). DASS trains the model by jointly using source domain data and some labeled or unlabeled target domain data so that the model can quickly generalize well to the target domain. Recent works can be grouped into adversarial training and self-training approaches. The former~\cite{Hoffman2017CyCADACA,Long2017ConditionalAD} aims to align the distributions of the source domain and the target domain via generative adversarial networks or Fourier transforms. The latter~\cite{Zou2018DomainAF,Zou2018UnsupervisedDA} is trained with pseudo-labels for the target domain. DGSS tries to bridge the domain gap through two main approaches, including Normalization and Whitening (NW), and Domain Randomization (DR). NW~\cite{Pan2019SwitchableWF,Peng2022SemanticAwareDG} normalizes the mean and standard deviation of the source data and whitens the covariance of the source data. DR~\cite{Peng2021GlobalAL,Huang2021FSDRFS} transforms source images into randomly stylized images and trains the network with them together. Unlike the setting of cross-domain semantic segmentation, CD-FSS not only has no access to target domain data during training, but also has disjoint label space between the source and target domains.\\
\noindent\textbf{Cross-domain Few-shot Semantic Segmentation.} Different from FSS, there are domain gaps between the source dataset and the target dataset, \emph{i.e.}, both data feature distributions and label spaces in the meta-testing stage are different from the meta-training stage. Recently, some works have been proposed to address this task. For example, PixDA~\cite{Tavera2021PixelbyPixelCA} proposes a pixel-by-pixel adversarial training strategy that uses a novel pixel-wise loss and discriminator to bridge the domain shift. To improve the generalization of the segmentation model, RTD~\cite{Wang2022RememberTD} designs a novel meta-memory module that transfers the intra-domain style information from the source domain images into target domain images. ~\cite{Lu2022CrossdomainFS} introduces a transductive fine-tuning method, which addresses the domain gap by using support labels to implicitly supervise query segmentation. PATNet~\cite{Lei2022CrossDomainFS} establishes a new evaluation benchmark for CD-FSS and converts the domain-specific features to domain-agnostic ones to enhance generalization ability. Our proposed method tackles several key issues, including overfitting by only utilizing support information for feature transformation, intra-class appearance variances, and under-utilized information, which are overlooked by previous works.

\section{Method}
\subsection{Problem Setting}
The problem setting of CD-FSS can be formulated as follows. There is a source domain $\left(\mathcal{X}_s,\mathcal{Y}_s\right)$ and a target domain $\left(\mathcal{X}_t,\mathcal{Y}_t\right)$. $\mathcal{X}_s$ and $\mathcal{X}_t$ represent the input data distributions while $\mathcal{Y}_s$ and $\mathcal{Y}_t$ represent the label spaces. In CD-FSS, $\mathcal{X}_s\ne\mathcal{X}_t$ and $\mathcal{Y}_s\cap\mathcal{Y}_t = \emptyset$, \emph{i.e.}, the input data distribution of the source domain is different from that of the target domain, and the label spaces of the domains do not intersect. Based on the data distributions and the label spaces, we construct the training set $\mathcal{D}_{train}$ and testing set $\mathcal{D}_{test}$. In term of $\emph{N}$-way $\emph{K}$-shot segmentation, both $\mathcal{D}_{train}$ and $\mathcal{D}_{test}$ consist of a large number of episodes. Each episode contains a support set $\mathcal{S} = \left \{ \left ( \mathcal{I}_{i}^{s},\mathcal{M}_{i}^{s}   \right )  \right \} _{i=1}^{N \times K}$ and a query set $\mathcal{Q} = \left \{ \left ( \mathcal{I}_{i}^{q},\mathcal{M}_{i}^{q}   \right )  \right \} _{i=1}^{Q}$, where $\mathcal{I}\in \mathbb{R}^{H\times W\times 3}$ denotes the RGB image and $\mathcal{M}\in \mathbb{R}^{H\times W}$ is the binary mask. So $\mathcal{D}_{train} = \left \{ \mathcal{I_{S/Q}}, \mathcal{M_{S/Q}} \right \}_{source} $ and $\mathcal{D}_{test} = \left \{ \mathcal{I_{S/Q}}, \mathcal{M_{S/Q}} \right \}_{target} $. In meta-training stage, the model is trained on $\mathcal{D}_{train}$, without exposure to the target domain. After completing episodes training, the model segmentation performance is evaluated using $\mathcal{D}_{test}$ in the meta-testing stage.
\subsection{Overview of DMTNet}
The overall architecture of DMTNet is illustrated in Figure ~\ref{fig2}. DMTNet consists of two major functional modules to bridge the domain gap: the Self-Matching Transformation (SMT) module and the Dual Hypercorrelation Construction (DHC) module.

During the meta-training stage, the support set and query set are first fed to a shared convolutional neural network to extract multi-level pyramid features. 
Then, we exploit SMT to learn each support and query image a transformation matrix, and transform the domain-specific features into domain-agnostic ones. 
Subsequently, DHC constructs the support foreground-query and support background-query hypercorrelations based on the domain-agnostic features, respectively. Finally, two available modules, 4D convolutional pyramid encoder and 2D convolutional pyramid decoder~\cite{Min2021HypercorrelationSF} are adopted to obtain the predicted query mask based on the dual hypercorrelations.

During the meta-testing stage, two steps are involved in obtaining the final prediction mask. We design a novel Test-time Self-Finetuning (TSF) strategy in the first step, where only fine-tuning a few parameters of the network can significantly refine the coarse mask in the second step and encourage the model to adapt quickly to the target domain.

\subsection{Self-Matching Transformation}
In cross-domain scenarios, domain style information in features hinders the model from accurately segmenting foreground objects. Therefore, ensuring the invariance of foreground objects while generalizing the domain style information can improve the generalization of representations. One useful strategy is to perform a feature transformation to transform the domain-relevant features into domain-irrelevant features. However, existing methods~\cite{Lei2022CrossDomainFS} transform the support and query image features only based on the prototypes of the support image, which may pose potential issues for meta-testing. 
As the support set during the meta-testing stage has only a few images for each class, repeatedly utilizing the same support images may cause overfitting. In addition, according to~\cite{Fan2022SelfSupportFS}, there may exist a huge appearance difference between the support and query images, even if they belong to the same class. In this case, the transformation matrix derived only from the support image may not be useful for the query images. 
This motivates us to reduce the dependence on the support features and mine information from the query image itself during feature transformation.
So we propose a Self-Matching Transformation (SMT) module. As shown in Figure~\ref{fig2}, SMT consists of two stages. In the first stage, a Similarity-based Self-matching module is used to generate a rough segmentation mask, based on which the prototype of the query image is obtained.
In the second stage, some learnable anchor layers are used to transform the support and query images' domain-specific features into domain-agnostic ones. 

\textbf{Similarity-based Self-matching.} Inspired by~\cite{Fan2022SelfSupportFS}, we first generate coarse segmentation masks for the query images via similarity-based self-matching between the query features with the foreground and background prototypes of the support image. Since compressing a global feature map into a prototype vector will lose much detailed information, we propose to divide the support features into several local features and generate a more fine-grained predicted query mask by measuring the similarity between support local prototypes and query global features.

Specifically, for a 1-way 1-shot task, we first obtain the $\mathit{L}$-level pyramid features of the support and query images, \emph{i.e.}, $\left \{ F_{l}^{s}, F_{l}^{q}  \right \} _{l=1}^{L}$, where $F_{l}\in \mathbb{R}^{C_{l}\times H_{l}\times W_{l}}$. Then we calculate the support global foreground prototype $\boldsymbol{c}_{fl}^{s}\in \mathbb{R}^{C_{l}}$ at intermediate layer $l$ with the support mask $\mathcal{M}^{s}\in  \left \{ 0, 1 \right \}^{H\times W}$ via Masked Average Pooling (MAP), as follows:
\begin{equation}
\boldsymbol{c}_{fl}^{s}=\frac{ {\textstyle \sum_{(x,y)}} F_{l}^{s}(x,y)\delta_{l}\left [ \mathcal{M}^{s} \right ](x,y)}{{\textstyle \sum_{(x,y)}\delta_{l}\left [ \mathcal{M}^{s} \right ](x,y)}},
\end{equation}
where $(x,y)$ are spatial positions, and $\delta_{l}\left [ \cdot \right ]$ denotes the bilinear interpolation. For simplicity, we represent $\delta_{l}\left [ \mathcal{M}^{s} \right ]$ as $\mathcal{M}_{l}^{s}$. Similarly, the support global background prototype $\boldsymbol{c}_{bl}^{s}$ can be calculated in the same way. Then we divide the support feature map $F_{l}^{s}$ into $\mathbf{P}$ support local feature maps, \emph{i.e.}, $\left \{ F_{lp}^{s} \right \}_{p=1}^{\mathbf{P} } $, $F_{lp}^{s}\in \mathbb{R}^{C_{l}\times \gamma H_{l}\times \gamma W_{l}}$. $\gamma$ is the division ratio and we set it to 0.25. We also obtain the local support masks $\left \{ \mathcal{M}_{lp}^{s} \right \}_{p=1}^{\mathbf{P} } $ following the same division. Then we calculate the $p$-th support local foreground prototype $\boldsymbol{c}_{flp}^{s}\in \mathbb{R}^{C_{l}}$ by MAP:
\begin{equation}
\boldsymbol{c}_{flp}^{s}=\frac{ {\textstyle \sum_{(x,y)}} F_{lp}^{s}(x,y)\mathcal{M}_{lp}^{s}(x,y)}{{\textstyle \sum_{(x,y)}\mathcal{M}_{lp}^{s}(x,y)}}.
\end{equation}
The p-th support local background prototype $\boldsymbol{c}_{blp}^{s}$ is similar to this. We then calculate the confidence matching correlation maps between query features and support local prototypes to obtain the naive query mask:
\begin{equation}
    \mathcal{M}_{1l}= \frac{1}{\mathbf{P} } {\textstyle \sum_{p=1}^{\mathbf{P} }} \left [ \eta \left ( \xi \left ( \boldsymbol{c}_{flp}^{s}, F_{l}^{q} \right )  \right ),\eta \left ( \xi \left ( \boldsymbol{c}_{blp}^{s}, F_{l}^{q} \right )  \right )  \right ],
\end{equation}
where $\eta\left( \cdot \right)$ denotes the softmax function. $\xi\left( \cdot \right)$ denotes a similarity measure function. In this work, we use the cosine similarity. With $\mathcal{M}_{1l}$, the query foreground and background prototype $\boldsymbol{c}_{fl}^{q}$, $\boldsymbol{c}_{bl}^{q}$ can be calculated.

To make the predicted rough query mask as accurate as possible, and thus 
provide a most accurate transformation matrix for later adaptive feature transformation, we propose to use a binary cross-entropy (BCE) loss function as supervision here, formulated as:
\begin{equation}
    \mathcal{L}_{1}= \frac{1}{L} {\textstyle \sum_{l=1}^{L}} \mathrm{BCE}\left ( \mathcal{M}_{1l}, \delta _{l}\left [ \mathcal{M}^{q} \right ] \right ).
\end{equation}

\textbf{Adaptive Feature Transformation.}
Similar to ~\cite{Seo2020TaskAdaptiveFT,Lei2022CrossDomainFS}, we use a linear transformation as the transformation mapper. The difference is that we construct specialized transformation matrices for support and query features, respectively, to ensure the invariance of the foreground objects during the adaptive transformation process. We construct the support prototype matrix and the query prototype matrix as $\mathbf{C}_{l}^{s}=\left [ \frac{\mathbf{c}_{fl}^{s}} {\left \| \mathbf{c} _{fl}^{s} \right \| } , \frac{\mathbf{c}_{bl}^{s}} {\left \| \mathbf{c} _{bl}^{s} \right \| }\right ] $, $\mathbf{C}_{l}^{q}=\left [ \frac{\mathbf{c}_{fl}^{q}} {\left \| \mathbf{c} _{fl}^{q} \right \| } , \frac{\mathbf{c}_{bl}^{q}} {\left \| \mathbf{c} _{bl}^{q} \right \| }\right ] $, respectively. Similarly, two trainable anchor weight matrices are defined as $\mathbf{A}_{l}^{s}=\left [ \frac{\mathbf{a}_{fl}^{s}} {\left \| \mathbf{a} _{fl}^{s} \right \| } , \frac{\mathbf{a}_{bl}^{s}} {\left \| \mathbf{a} _{bl}^{s} \right \| }\right ] $, $\mathbf{A}_{l}^{q}=\left [ \frac{\mathbf{a}_{fl}^{q}} {\left \| \mathbf{a} _{fl}^{q} \right \| } , \frac{\mathbf{a}_{bl}^{q}} {\left \| \mathbf{a} _{bl}^{q} \right \| }\right ]$, where $\mathbf{a}_{fl}^{s},\mathbf{a}_{bl}^{s},\mathbf{a}_{fl}^{q},\mathbf{a}_{bl}^{q}\in \mathbb{R}^{C_{l}} $. Similar to PATNet~\cite{Lei2022CrossDomainFS}, we only set three anchor layers for support and query features, respectively. They correspond to the feature maps of three dimensions, \emph{i.e.}, low, medium and high-level features. So we construct two transformation matrices $\left [ \mathbf{W}_{l}^{s},\mathbf{W}_{l}^{q} \right ]$ by solving $\mathbf{W}_{l}^{s}\mathbf{C}_{l}^{s}=\mathbf{A}_{l}^{s} $ and $\mathbf{W}_{l}^{q}\mathbf{C}_{l}^{q}=\mathbf{A}_{l}^{q} $. Since $\mathbf{C}_{l}^{s}$ and $\mathbf{C}_{l}^{q}$ are non-square matrices, we calculate their generalized inverse by $\mathbf{C}_{l}^{s+}=\left \{ \mathbf{C}_{l}^{sT} \mathbf{C}_{l}^{s}  \right \}^{-1}\mathbf{C}_{l}^{sT}$, $\mathbf{C}_{l}^{q+}=\left \{ \mathbf{C}_{l}^{qT} \mathbf{C}_{l}^{q}  \right \}^{-1}\mathbf{C}_{l}^{qT}$. In particular, we propose to refine $\mathbf{C}_{l}^{q+}$ by integrating $\mathbf{C}_{l}^{s+}$ into $\mathbf{C}_{l}^{q+}$:
\begin{equation}
    \mathbf{C}_{l}^{q+} = \beta  \mathbf{C}_{l}^{q+} + \left ( 1-\beta  \right ) \mathbf{C}_{l}^{s+},
\end{equation}
where $\beta$ mainly controls the integration ratio, which is set to 0.5. Finally, the support and query transformation matrices at layer $l$ can be calculated as $\mathbf{W}_{l}^{s}=\mathbf{A}_{l}^{s}\mathbf{C}_{l}^{s+}$, $\mathbf{W}_{l}^{q}=\mathbf{A}_{l}^{q}\mathbf{C}_{l}^{q+}$ respectively, where $\mathbf{W}_{l}^{s}, \mathbf{W}_{l}^{q}\in \mathbb{R}^{C^{l}\times C^{l}}$.

\subsection{Dual Hypercorrelation Construction}
Considering that objects from the same category are very likely to lie in similar environments, the background information correlations between the query and support images can also be used in CD-FSS. 
However, the existing method~\cite{Min2021HypercorrelationSF} directly filters out the support background region using the support mask. 
Differently, we propose a Dual Hypercorrelation Construction module (DHC) to explore the dense correlations between the query features with the foreground and background features of the support images in domain-agnostic space.  

Firstly, we construct the 4D correlation tensor $\mathbf{Corr}_{fl}\in \mathbb{R}^{H_{l}\times W_{l} \times H_{l}\times W_{l}} $ based on the support foreground features $F_{fl}^{s}$ and query features $F_{l}^{q}$ by cosine similarity:
\begin{equation}
    \mathbf{Corr}_{fl}=\mathrm{ReLU}\left ( \frac{\mathbf{W}_{l}^{s}F_{fl}^{s}(x_{1},y_{1})\cdot \mathbf{W}_{l}^{q}F_{l}^{q}(x_{2},y_{2}) }{\left \| \mathbf{W}_{l}^{s}F_{fl}^{s}(x_{1},y_{1}) \right \| \cdot \left \| \mathbf{W}_{l}^{q}F_{l}^{q}(x_{2},y_{2}) \right \|  }  \right ),
\end{equation}
where $(x_{1},y_{1})$ and $(x_{2},y_{2})$ denotes 2D spatial positions of support and query feature maps, respectively.

Secondly, we construct $\mathbf{Corr}_{bl}$ based on the support background features $F_{bl}^{s}$ and query features $F_{l}^{q}$:
\begin{equation}
    \mathbf{Corr}_{bl}=\mathrm{ReLU}\left ( \frac{\mathbf{W}_{l}^{s}F_{bl}^{s}(x_{1},y_{1})\cdot \mathbf{W}_{l}^{q}F_{l}^{q}(x_{2},y_{2}) }{\left \| \mathbf{W}_{l}^{s}F_{bl}^{s}(x_{1},y_{1}) \right \| \cdot \left \| \mathbf{W}_{l}^{q}F_{l}^{q}(x_{2},y_{2}) \right \|  }  \right ). 
\end{equation}

Then, the dense correlation maps are fed to two available modules, 4D convolutional pyramid encoder and 2D convolutional pyramid decoder~\cite{Min2021HypercorrelationSF} to generate the predicted query foreground mask $\mathcal{M}_{f}$ and background mask $\mathcal{M}_{b}$. The training supervision on the two predicted masks is:
\begin{equation}
    \mathcal{L}_{2}=\mathrm{BCE}\left ( \mathcal{M}_{f}, \mathcal{M}^{q} \right ) + \alpha_{1} \cdot \mathrm{BCE}\left ( \mathcal{M}_{b}, 1-\mathcal{M}^{q} \right ),
\end{equation}
where $\alpha_{1}$ is a tuning weight. Finally, we train the model in an end-to-end manner by jointly optimizing all the losses:
\begin{equation}
    \mathcal{L}=\alpha_{2} \cdot \mathcal{L}_{1}+\mathcal{L}_{2},
\end{equation}
where $\alpha_{2}$ is a balancing hyperparameter. $\alpha_{1}$ and $\alpha_{2}$ are set as 1.0 and 0.5, respectively.

\subsection{Test-time Self-Finetuning}
In the meta-testing stage, we design a Test-time Self-Finetuning (TSF) strategy to refine query predictions in unseen domains. PATNet\cite{Lei2022CrossDomainFS} proposes to fine-tune the anchor layers by reducing the distribution distance between support foreground prototypes and the query foreground prototypes which are obtained by the predicted query mask. However, they assume images belonging to the same class have similar appearances, which is not always true in few-shot scenarios. We contend that there may exist significant appearance variances even within the same class. Therefore, aligning the foreground prototypes of the support and query images may cause overfitting to support images and distortion of query foreground features. To this end, we propose TSF to self-tune the network by trying to predict the ground-truth masks of the support images. By finetuning the network on support images, our model can learn style information of the target domain, leading to the generation of more accurate masks for the query images.

As shown in the Figure~\ref{fig2}, TSF has two steps. In the first step, the model outputs the predicted support masks $\overline{\mathcal{M}^{s}}$ and updates the network with the loss:
\begin{equation}
    \mathcal{L}_{TSF}=\frac{1}{K} {\textstyle \sum_{k=1}^{K}}\mathrm{BCE}\left ( \overline{\mathcal{M}_{k}^{s}}, \mathcal{M}_{k}^{s}\right ).
\end{equation}
In the second step, we freeze the entire network and execute the final prediction for the query image.

Similar to PATNet, we do not finetune the whole network. Instead, we only finetune a few parameters of the encoder, which is validated by the 
quantitative experiments. The details can be found in Section 4.4.

\section{Experiments}
\begin{table*}[]
\centering
\resizebox{1.0\linewidth}{!}{
\begin{tabular}{cccccccccccc}
\hline
\multicolumn{1}{c|}{\multirow{2}{*}{Methods}} & \multicolumn{1}{c|}{\multirow{2}{*}{Backbone}} & \multicolumn{2}{c|}{ISIC}                            & \multicolumn{2}{c|}{Chest X-ray}                     & \multicolumn{2}{c|}{Deepglobe}                       & \multicolumn{2}{c|}{FSS-1000}                        & \multicolumn{2}{c}{Average}     \\ \cline{3-12} 
\multicolumn{1}{c|}{}                         & \multicolumn{1}{c|}{}                          & 1-shot         & \multicolumn{1}{c|}{5-shot}         & 1-shot         & \multicolumn{1}{c|}{5-shot}         & 1-shot         & \multicolumn{1}{c|}{5-shot}         & 1-shot         & \multicolumn{1}{c|}{5-shot}         & 1-shot         & 5-shot         \\ \hline
\multicolumn{12}{c}{Transfer Learning Methods}                                                                                                                                                                                                                                                                                                               \\ \hline
\multicolumn{1}{c|}{Ft-last-$1_{\mathrm{FCN}}$}             & \multicolumn{1}{c|}{VGG-16}                    & 15.17          & \multicolumn{1}{c|}{19.75}          & 33.63          & \multicolumn{1}{c|}{48.08}          & 29.80          & \multicolumn{1}{c|}{32.25}          & 32.51          & \multicolumn{1}{c|}{53.62}          & 27.78          & 38.43          \\
\multicolumn{1}{c|}{Ft-last-$2_{\mathrm{FCN}}$}             & \multicolumn{1}{c|}{VGG-16}                    & 17.52          & \multicolumn{1}{c|}{21.65}          & 36.35          & \multicolumn{1}{c|}{53.85}          & 32.90          & \multicolumn{1}{c|}{35.34}          & 32.15          & \multicolumn{1}{c|}{57.44}          & 29.82          & 42.07          \\
\multicolumn{1}{c|}{Ft-last-$3_{\mathrm{FCN}}$}             & \multicolumn{1}{c|}{VGG-16}                    & 17.91          & \multicolumn{1}{c|}{25.58}          & 45.61          & \multicolumn{1}{c|}{56.05}          & 32.91          & \multicolumn{1}{c|}{35.54}          & 33.32          & \multicolumn{1}{c|}{{60.86}}    & 32.34          & 44.51          \\
\multicolumn{1}{c|}{1N$\mathrm{N}_{FCN}$}                   & \multicolumn{1}{c|}{VGG-16}                    & 15.68          & \multicolumn{1}{c|}{23.66}          & 46.26          & \multicolumn{1}{c|}{52.70}          & 32.42          & \multicolumn{1}{c|}{38.63}          & 41.51          & \multicolumn{1}{c|}{46.64}          & 33.97          & 40.41          \\
\multicolumn{1}{c|}{Linea$\mathrm{r}_{FCN}$}                & \multicolumn{1}{c|}{VGG-16}                    & 15.51          & \multicolumn{1}{c|}{{30.65}}    & 37.69          & \multicolumn{1}{c|}{50.07}          & {33.56}    & \multicolumn{1}{c|}{38.75}          & 41.09          & \multicolumn{1}{c|}{49.16}          & 31.96          & 42.16          \\
\multicolumn{1}{c|}{Ft-last-$1_{Deeplab}$}         & \multicolumn{1}{c|}{ResNet-50}                    & 11.08          & \multicolumn{1}{c|}{16.57}          & 30.43          & \multicolumn{1}{c|}{35.54}          & 28.11          & \multicolumn{1}{c|}{28.65}          & 25.14          & \multicolumn{1}{c|}{35.86}          & 23.69          & 29.41          \\
\multicolumn{1}{c|}{Ft-last-$2_{Deeplab}$}         & \multicolumn{1}{c|}{ResNet-50}                    & 10.22          & \multicolumn{1}{c|}{17.56}          & 31.16          & \multicolumn{1}{c|}{51.57}          & 24.09          & \multicolumn{1}{c|}{36.74}          & 20.68          & \multicolumn{1}{c|}{42.50}          & 21.29          & 37.10          \\
\multicolumn{1}{c|}{1N-$N_{Deeplab}$}               & \multicolumn{1}{c|}{ResNet-50}                    & {21.44}     & \multicolumn{1}{c|}{26.04}          & {47.76}    & \multicolumn{1}{c|}{57.93}          & 32.28          & \multicolumn{1}{c|}{35.96}          & {45.81}    & \multicolumn{1}{c|}{55.95}          & {36.82}    & 43.97          \\
\multicolumn{1}{c|}{Linea$\mathrm{r}_{Deeplab}$}            & \multicolumn{1}{c|}{ResNet-50}                    & 19.42          & \multicolumn{1}{c|}{30.04}          & 43.52          & \multicolumn{1}{c|}{{60.29}}   & 32.95          & \multicolumn{1}{c|}{{39.69}}    & 40.50          & \multicolumn{1}{c|}{58.36}          & 34.10          & {47.10}    \\ \hline
\multicolumn{12}{c}{Few-shot Semantic Segmentation Methods}                                                                                                                                                                                                                                                                                                  \\ \hline
\multicolumn{1}{c|}{AMP~\cite{Siam2019AMPAM}}                      & \multicolumn{1}{c|}{VGG-16}                    & 28.42          & \multicolumn{1}{c|}{30.41}          & 51.23          & \multicolumn{1}{c|}{53.04}          & 37.61          & \multicolumn{1}{c|}{40.61}          & 57.18          & \multicolumn{1}{c|}{59.24}          & 43.61          & 45.83          \\
\multicolumn{1}{c|}{PGNet~\cite{Zhang2019PyramidGN}}                    & \multicolumn{1}{c|}{ResNet-50}                    & 21.86          & \multicolumn{1}{c|}{21.25}          & 33.95          & \multicolumn{1}{c|}{27.96}          & 10.73          & \multicolumn{1}{c|}{12.36}          & 62.42          & \multicolumn{1}{c|}{62.74}          & 32.24          & 31.08          \\
\multicolumn{1}{c|}{PANet~\cite{Wang2019PANetFI}}                    & \multicolumn{1}{c|}{ResNet-50}                    & 25.29          & \multicolumn{1}{c|}{33.99}          & 57.75          & \multicolumn{1}{c|}{69.31}          & 36.55          & \multicolumn{1}{c|}{45.43}          & 69.15          & \multicolumn{1}{c|}{71.68}          & 47.19          & 55.10          \\
\multicolumn{1}{c|}{CaNet~\cite{Zhang2019CANetCS}}                    & \multicolumn{1}{c|}{ResNet-50}                    & 25.16          & \multicolumn{1}{c|}{28.22}          & 28.35          & \multicolumn{1}{c|}{28.62}          & 22.32          & \multicolumn{1}{c|}{23.07}          & 70.67          & \multicolumn{1}{c|}{72.03}          & 36.63          & 37.99          \\
\multicolumn{1}{c|}{RPMMs~\cite{Yang2020PrototypeMM}}                    & \multicolumn{1}{c|}{ResNet-50}                    & 18.02          & \multicolumn{1}{c|}{20.04}          & 30.11          & \multicolumn{1}{c|}{30.82}          & 12.99          & \multicolumn{1}{c|}{13.47}          & 65.12          & \multicolumn{1}{c|}{67.06}          & 31.56          & 32.85          \\
\multicolumn{1}{c|}{PFENet~\cite{Tian2020PriorGF}}                   & \multicolumn{1}{c|}{ResNet-50}                    & 23.50          & \multicolumn{1}{c|}{23.83}          & 27.22          & \multicolumn{1}{c|}{27.57}          & 16.88          & \multicolumn{1}{c|}{18.01}          & 70.87          & \multicolumn{1}{c|}{70.52}          & 34.62          & 34.98          \\
\multicolumn{1}{c|}{RePRI~\cite{Boudiaf2020FewShotSW}}                    & \multicolumn{1}{c|}{ResNet-50}                    & 23.27          & \multicolumn{1}{c|}{26.23}          & 65.08          & \multicolumn{1}{c|}{65.48}          & 25.03          & \multicolumn{1}{c|}{27.41}          & 70.96          & \multicolumn{1}{c|}{74.23}          & 46.09          & 48.34          \\
\multicolumn{1}{c|}{HSNet~\cite{Min2021HypercorrelationSF}}                    & \multicolumn{1}{c|}{ResNet-50}                    & 31.20          & \multicolumn{1}{c|}{35.10}          & 51.88          & \multicolumn{1}{c|}{54.36}          & 29.65          & \multicolumn{1}{c|}{35.08}          & 77.53          & \multicolumn{1}{c|}{80.99}          & 47.57          & 51.38          \\ \hline
\multicolumn{12}{c}{Cross-domain Few-shot Semantic Segmentation Methods}                                                                                                                                                                                                                                                         \\ \hline
\multicolumn{1}{c|}{PATNet~\cite{Lei2022CrossDomainFS}}                   & \multicolumn{1}{c|}{VGG-16}                    & 33.07       & \multicolumn{1}{c|}{45.83}       & 57.83       & \multicolumn{1}{c|}{60.55}       & 28.74       & \multicolumn{1}{c|}{34.83}       & 71.60       & \multicolumn{1}{c|}{76.17}       & 47.81        & 54.35        \\
\multicolumn{1}{c|}{PATNet~\cite{Lei2022CrossDomainFS}}                   & \multicolumn{1}{c|}{ResNet-50}                    & {41.16}    & \multicolumn{1}{c|}{\textbf{53.58}} & 66.61          & \multicolumn{1}{c|}{70.20}          & {\underline {37.89}}    & \multicolumn{1}{c|}{42.97}          & {78.59}    & \multicolumn{1}{c|}{{81.23}}    & {\underline{56.06}}    & {\underline{61.99}}    \\
\multicolumn{1}{c|}{RestNet~\cite{huang2023restnet}}                   & \multicolumn{1}{c|}{ResNet-50}                    & {\underline{42.25}}    & \multicolumn{1}{c|}{51.10} & 70.43          & \multicolumn{1}{c|}{73.69}          & {22.70}    & \multicolumn{1}{c|}{29.99}          & {\underline {81.53}}    & \multicolumn{1}{c|}{{\underline {84.89}}}    & {54.23}    & { 59.92}    \\ 
\multicolumn{1}{c|}{DAM~\cite{chen2023dense}}                   & \multicolumn{1}{c|}{ResNet-50}                    & {-}    & \multicolumn{1}{c|}{-} & 70.4          & \multicolumn{1}{c|}{74.0}          & {37.1}    & \multicolumn{1}{c|}{41.6}          & {\textbf {84.6}}    & \multicolumn{1}{c|}{{\textbf {86.3}}}    & {-}    & {-}    \\ \hline
\multicolumn{1}{c|}{\textbf{DMTNet}}    & \multicolumn{1}{c|}{VGG-16}                    & 34.26          & \multicolumn{1}{c|}{40.66}          & {\underline {73.02}}    & \multicolumn{1}{c|}{{\underline {75.84}}}    & 34.85          & \multicolumn{1}{c|}{{\underline {47.77}}}    & 74.32          & \multicolumn{1}{c|}{77.11}          & 54.11          & 60.35          \\
\multicolumn{1}{c|}{\textbf{DMTNet}}    & \multicolumn{1}{c|}{ResNet-50}                    & \textbf{43.55} & \multicolumn{1}{c|}{\underline{52.30}}    & \textbf{73.74} & \multicolumn{1}{c|}{\textbf{77.30}} & \textbf{40.14} & \multicolumn{1}{c|}{\textbf{51.17}} & 81.52 & \multicolumn{1}{c|}{83.28} & \textbf{59.74} & \textbf{66.01} \\ \hline
\end{tabular}
}
\caption{Performance of transfer learning, FSS, and CD-FSS methods in Mean-IoU under (1-way) 1-shot and (1-way) 5-shot settings. The best and second-best results are in bold and underlined, respectively.}
\label{mainresult}
\end{table*}
\begin{table}[]
\centering
\begin{tabular}{cccc|cc}
\hline
      & SMT     & DHC     & TSF     & 1-shot                          & $\bigtriangleup$   \\ \hline
DMTNet & $\surd$ & $\surd$ & $\surd$ &  \textbf{59.74} & 0.0                \\
-TSF  & $\surd$ & $\surd$ &         & 57.23                           & $\downarrow$ 2.51  \\
-DHC  & $\surd$ &         &         & 56.36                           & $\downarrow$ 3.35  \\
-STM  &         &         &         & 47.57                           & $\downarrow$ 12.17 \\ \hline
\end{tabular}
\caption{Ablation results of our proposed modules with 1-shot performance averaged over four datasets.}
\label{ablation1}
\end{table}
\begin{figure}[ht]
    \centering
    \includegraphics[scale=0.3]{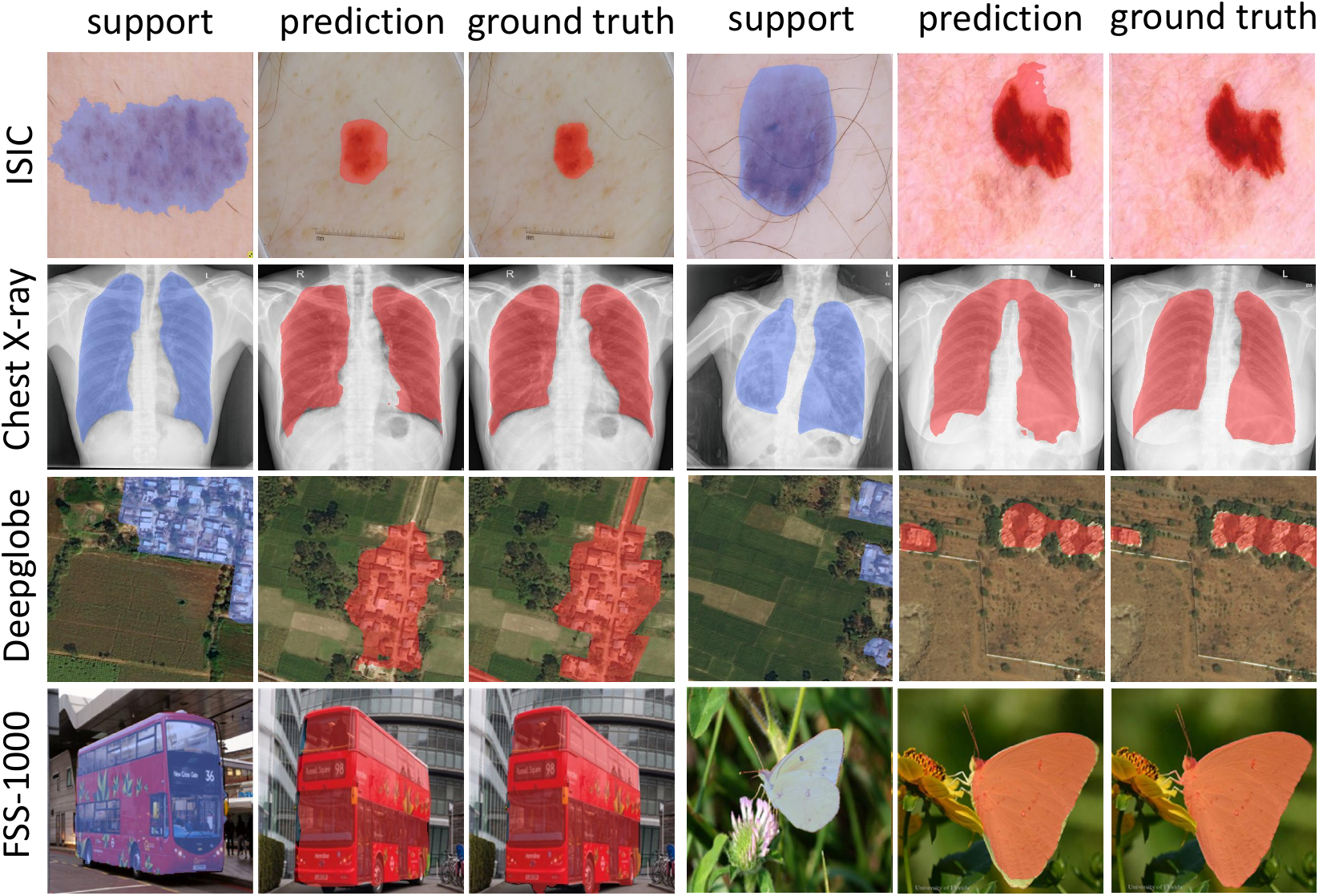}
    \caption{Qualitative results on the ISIC, Chest X-ray, Deepglobe, and FSS-1000 datasets under the 1-shot setting. The blue parts represent support masks and the red parts represent query masks and query predictions.}
    \label{fig3}
\end{figure}
\begin{figure}[ht]
    \centering
    \includegraphics[scale=0.28]{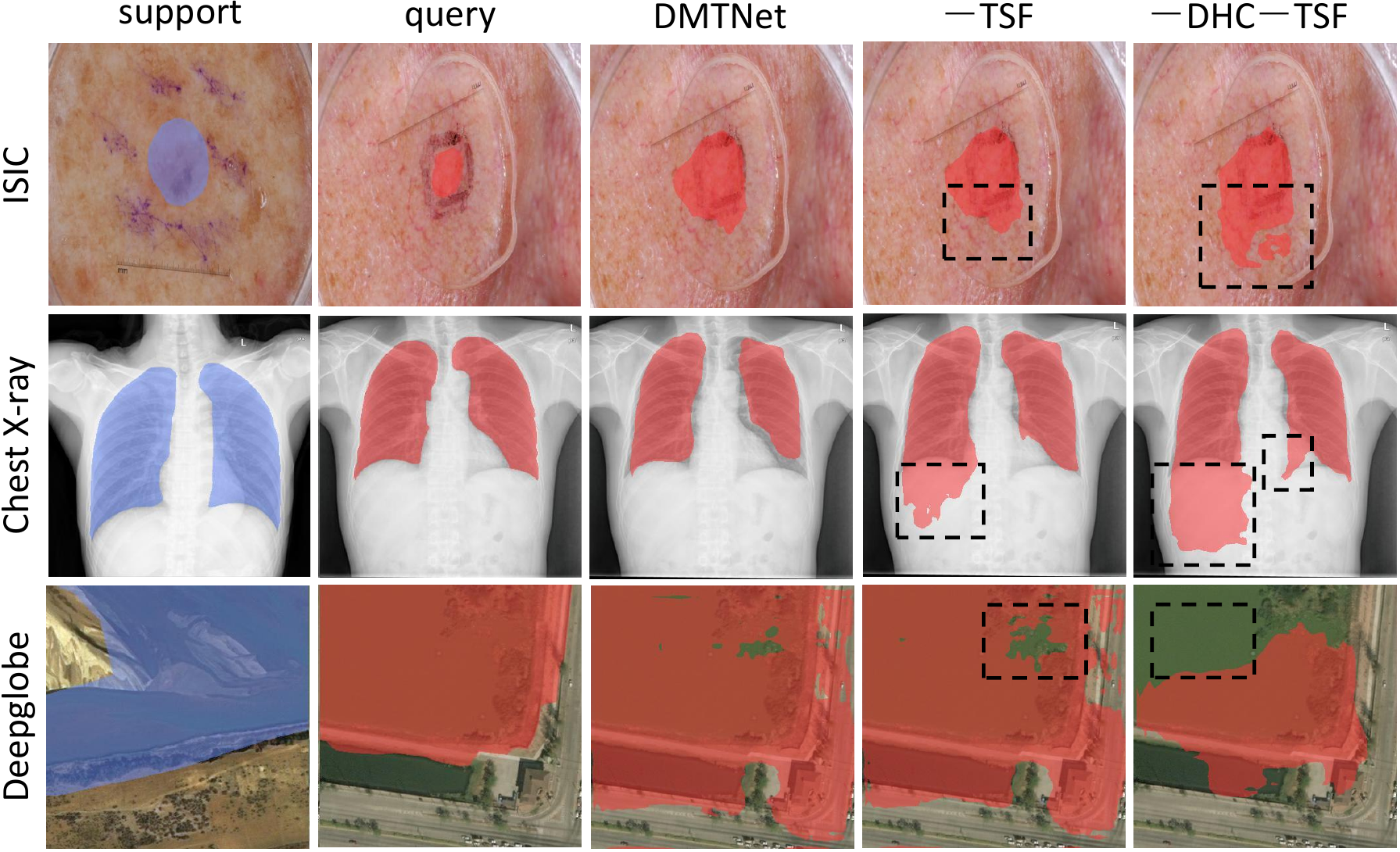}
    \caption{Qualitative results w.r.t. SMT, DHC, and TSF. The first two columns show the ground truth of the support and query images. The third column shows the predicted masks of DMTNet. The fourth column shows the prediction masks without the TSF. The last column shows the prediction masks without DHC and TSF modules.}
    \label{fig4}
\end{figure}

\subsection{Experimental Setup}
To fairly compare the cross-domain segmentation performance of DMTNet with PATNet~\cite{Lei2022CrossDomainFS}, we choose PASCAL VOC 2012 with SBD augmentation as the source domain, and ISIC2018, Chest X-ray, Deepglobe, and FSS-1000 as the target domains.

\textbf{ISIC2018} is a skin cancer screening dataset consisting of lesion images. The dataset contains three types of skin lesions and a total of 2,596 images, each with one primary lesion region. The initial spatial resolution is approximately 1022 $\times$ 767 and we uniformly reduce it to 512 $\times$ 512.

\textbf{Chest X-ray} is a Tuberculosis X-ray image dataset with 566 annotated images, which is collected from 58 abnormal cases with a manifestation of Tuberculosis and 80 normal cases. The initial spatial resolution is 4020 $\times$ 4892 and we downsize it to 1024 $\times$ 1024.

\textbf{Deepglobe} is a satellite image dataset. Each image is densely annotated at pixel level with 7 categories: urban land, agricultural land, rangeland, forestland, water, barren land, and unknown. Following PATNet, we partition each image into 6 pieces and filter the single class images and the `unknown' class, leading to a dataset of 5,666 images with a spatial resolution of 408 $\times$ 408. 

\textbf{FSS-1000} is a natural image dataset specialized for FSS, which contains 1,000 categories and each category has 10 annotated images. Each image contains only one segmentation target and has a resolution of 224 $\times$ 224. 

We use the standard Intersection over Union (IoU) metric by averaging the prediction results across 5 runs with different random seeds. Each run is composed of 1,200 tasks for ISIC2018, Chest X-ray, and Deepglobe. And for FSS-1000, each run contains 2,400 tasks. We evaluate our model under 1-way 1-shot and 1-way 5-shot settings.

\subsection{Implementation Details}
Following previous works, we employ ResNet-50 and VGG-16 pre-trained on ILSVRC as our backbones. For ResNet-50, the features from the conv3\_x, conv4\_x, and conv5\_x are extracted to produce feature maps. The channel dimensions of the three anchor layers are set to 512, 1024, and 2048, respectively. For the VGG-16 backbone, the features from the conv4\_x to conv5\_x are extracted to produce feature maps. The channel dimensions of the three anchor layers all are set to 512. In the meta-training stage, we use Adam optimizer to train DMTNet for 19 epochs with a learning rate of 1e-3. In the self-finetuning of the meta-testing stage, we use Adam optimizer with a learning rate of 1e-6 for ISIC2018, Deepglobe and FSS-1000, 1e-1 for Chest X-ray. 
All input images are resized to 400 $\times$ 400 resolution.
\subsection{Comparison with State-of-the-Art Methods}
We compare the performance of our model with several state-of-the-art methods. These methods are categorized into three groups: transfer learning, few-shot semantic segmentation, and cross-domain few-shot semantic segmentation methods. All methods are trained on PASCAL VOC and tested on the four datasets. The experimental results are shown in Table~\ref{mainresult}. Because the whole ISIC is seen as one class in DAM~\cite{chen2023dense}, we do not compare their report results on ISIC here (under the same ISIC dataset setting, our model exceeds DAM by 2.25\% in the 5-shot setting). We can see that under both (1-way) 1-shot and (1-way) 5-shot settings, the performance of our model ranks at the top on the average results of the four datasets, reaching 59.74\% MIoU in the 1-shot setting and 66.01\% MIoU in the 5-shot setting. We show the 1-way segmentation visualization results of our model on four datasets in Figure \ref{fig3}.

When compared with the state-of-the-art PATNet~\cite{Lei2022CrossDomainFS}, our model exceeds PATNet by 3.68\% and 4.02\% in the 1-shot and 5-shot settings respectively. This suggests that using query-specific transformation matrices and the dual hypercorrelation construction can avoid overfitting to base classes, leading to further enhancement of the segmentation results.
In addition, when using VGG-16 as the backbone, our model significantly improves performance, leading to the second-best results on Chest X-ray and Deepglobe.
\begin{table}[]
\centering
\setlength{\tabcolsep}{1mm}{
\begin{tabular}{ccccc|c}
\hline
$\mathrm{Ft}_{low}$  & $\mathrm{Ft}_{mid}$  & $\mathrm{Ft}_{high}$ & $\mathrm{Ft}_{encoder}$ & $\mathrm{Ft}_{decoder}$ & Average \\ \hline
$\surd$              &                      &                      &                         &                         & 57.69   \\
                     & $\surd$              &                      &                         &                         & 58.72        \\
                     &                      & $\surd$              &                         &                         & 58.95        \\
\multicolumn{1}{l}{} & \multicolumn{1}{l}{} & \multicolumn{1}{l}{} & $\surd$                 & \multicolumn{1}{l|}{}   & \textbf{59.74}   \\
                     &                      &                      &                         & $\surd$                 & 53.16        \\ \hline
\end{tabular}}
\caption{Ablation results on the choice of fine-tuning parameters for TSF under 1-way 1-shot setting.}
\label{ablation2}
\end{table}
\subsection{Ablation Study}

\textbf{Component Analysis.}
We conduct several ablation experiments to verify the effectiveness of the three key modules of DMTNet, \emph{i.e.} SMT, DHC and TSF. Table~\ref{ablation1} shows the impact of each module on the model performance. We can see that using all three modules proposed in this paper achieves the best results, and removing any of them would lead to a drop in the average performance across four datasets. Figure~\ref{fig4} further shows the segmentation visualization results of removing the TSF and the TSF+DHC modules, demonstrating the impact of our proposed module in a more intuitive way.
 \begin{figure}[ht]
    \centering
    \includegraphics[scale=0.32]{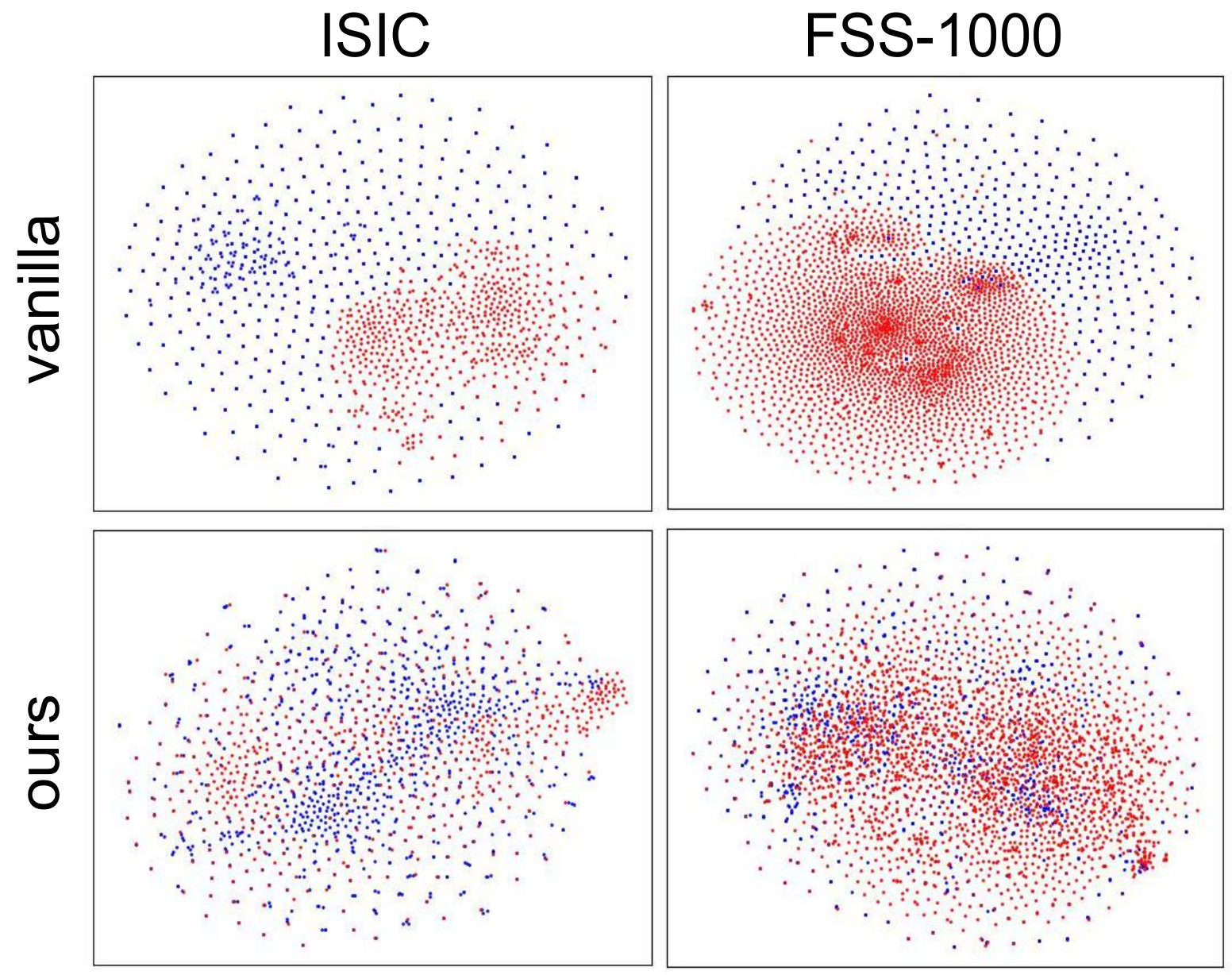}
    \caption{Visualization results w.r.t. SMT. The first and second rows represent the feature distributions before and after applying SMT, respectively. The red dots represent the PASCAL VOC dataset and the blue dots represent the ISIC or FSS-1000 datasets.}
    \label{fig5}
\end{figure}

\textbf{Fine-tuning Parameters for TSF.}
We conduct quantitative experiments to select the fine-tuning parameters for the test-time self-finetuning strategy. Following PATNet~\cite{Lei2022CrossDomainFS}, we choose the low-, medium- and high-level anchor layers. 
Additionally, we believe that the encoder/decoder should have an important impact on the cross-domain segmentation performance as it hierarchically fuses/reconstructs the 4D correlation maps, therefore we also choose the two fusing/de-fusing layers in the encoder/decoder. The results are presented in Table~\ref{ablation2}. We can observe that fine-tuning all but the decoder layers can lead to performance improvement, demonstrating the effectiveness of the TSF. Fine-tuning the encoder layers achieves the best performance of 59.45\%, indicating the effectiveness of bringing target domain information into the encoder for better fusion of correlation maps. 

\textbf{Visualization w.r.t. SMT.}
We visualize the image feature distributions before and after SMT to more clearly illustrate the effect of SMT in bridging the domain gaps, as shown in Figure~\ref{fig5}. We show the feature distributions of ISIC and FSS-1000 datasets for comparison with PASCAL VOC. We can see that the feature distributions of ISIC and FSS-1000 are closer to the PSCAL VOC after using SMT, which proves the effectiveness of our module in bridging domain distances and better generalization to the target domains.
\section{Conclusion}
We propose DMTNet for cross-domain few-shot semantic segmentation. DMTNet first exploits an SMT module to calculate each support and query image a transformation matrix based on its own prototype, and transform their domain-specific features into domain-agnostic ones self-adaptively. Then, a DHC module is used to explore the dual hypercorrelation between the query image with the foreground and background of the support image in the domain-agnostic space, based on which a foreground and background prediction mask are obtained and supervised during meta-training, respectively. During meta-testing, a TSF strategy is used to further improve the segmentation performance by only refining a few parameters of DMTNet. Extensive experiments show that DMTNet is effective and achieves state-of-the-art performance on four datasets with different domain gaps.
\section*{Acknowledgments}
This work was partially supported by the National Natural Science Foundation of China (No. 62276129 \& No. 62206127), and the Natural Science Foundation of Jiangsu Province (No. BK20220890).
\bibliographystyle{named}
\bibliography{ijcai24}
\end{document}